\documentclass{article}
\usepackage{xcolor}
\definecolor{blue3f}{HTML}{1a80bb}
\definecolor{orange3f}{HTML}{ea801c}
\definecolor{gray3f}{HTML}{b8b8b8}
\definecolor{red3f}{HTML}{a00000}
\usepackage{microtype}
\usepackage{graphicx}
\usepackage{subcaption}
\usepackage{booktabs} % for professional tables

\usepackage{hyperref}

\usepackage[preprint]{icml2026}

\usepackage{amsmath}
\usepackage{amssymb}
\usepackage{mathtools}
\usepackage{amsthm}
\usepackage{tikz}
\usetikzlibrary{positioning, arrows.meta, shapes}
\usepackage{pgfplots} 

\usepackage[capitalize,noabbrev]{cleveref}

\theoremstyle{plain}

\theoremstyle{definition}

\theoremstyle{remark}

\usepackage[textsize=tiny]{todonotes}

\begin{document}

\twocolumn[
  \icmltitle{Memories Retrieved from Many Paths: A Multi-Prefix Framework for Robust Detection of Training Data Leakage in Large Language Models}
  \icmltitlerunning{A Multi-Prefix Framework for Robust Detection of Training Data Leakage in Large Language Models}



  \icmlsetsymbol{equal}{*}

  \begin{icmlauthorlist}
    \icmlauthor{Trung Cuong Dang}{yyy}
    \icmlauthor{David Mohaisen}{yyy}
  \end{icmlauthorlist}

  \icmlaffiliation{yyy}{Department of Computer Science, University of Central Florida}
  
  \icmlcorrespondingauthor{Trung Cuong Dang}{cuong.dang@ucf.edu}
      \icmlcorrespondingauthor{David Mohaisen}{mohaisen@ucf.edu}

  \icmlkeywords{Machine Learning, ICML}

  \vskip 0.3in
]



\printAffiliationsAndNotice{}  % no special notice (required even if empty)

\begin{abstract}
Large language models, trained on massive corpora, are prone to verbatim memorization of training data, creating significant privacy and copyright risks. While previous works have proposed various definitions for memorization, many exhibit shortcomings in comprehensively capturing this phenomenon, especially in aligned models. To address this, we introduce a novel framework: multi-prefix memorization. Our core insight is that memorized sequences are deeply encoded and thus retrievable via a significantly larger number of distinct prefixes than non-memorized content. We formalize this by defining a sequence as memorized if an external adversarial search can identify a target count of distinct prefixes that elicit it. This framework shifts the focus from single-path extraction to quantifying the robustness of a memory, measured by the diversity of its retrieval paths. Through experiments on open-source and aligned chat models, we demonstrate that our multi-prefix definition reliably distinguishes memorized from non-memorized data, providing a robust and practical tool for auditing data leakage in LLMs.
\end{abstract}

\section{Introduction}
% Problem introduction
Large language models have demonstrated remarkable capabilities in tasks ranging from advanced text generation to nuanced reasoning and interactive dialogue. Their power stems from training on vast Internet-scale datasets, enabling them to internalize extensive knowledge and complex linguistic structures~\citep{BrownMRSKDNSSAA20}. However, this training paradigm introduces a critical challenge: the tendency of LLMs to memorize and reproduce segments of their training data verbatim. This phenomenon, which has been demonstrated to occur even in state-of-the-art models~\citep{CarliniTWJHLRBS21}, raises concerns regarding privacy, copyright, and the ethical deployment of these models. Consequently, establishing a precise definition for memorization is of paramount importance, with implications for both practical applications and legal frameworks~\citep{cooper2023genai}, and the existing definitions offer varied perspectives on memorization. 

Some initial formulations define memorization based on simple elicitation, where a sequence is considered memorized if any prompt can be found that reproduces it exactly~\citep{NasrRCHJCICTL25}. The weakness of this definition, however, is that it conflates true memorization with basic instruction-following, as a prompt could simply contain the target sequence itself (e.g., ``Repeat the following text: ..."). To tackle this, more constrained definitions have been proposed. A widely adopted version is {\em discoverable memorization}~\citep{CarliniIJLTZ23}, which defines the concept as a completion test using the target sequence's own prefix. This approach, however, lacks robustness against modern alignment techniques. Its reliance on natural prefixes makes it susceptible to evasion by models trained to refuse such completions, creating an ``illusion of compliance'' even when the memory persists~\citep{ippolito2022preventing}. 

To address this issue, an information-theoretic definitions has been introduced in~\cite{SchwarzschildFM24}, which defines that a sequence is memorized if it can be elicited by a prompt (significantly) shorter than the sequence itself. While this definition is more resilient to surface-level alignment, its practical utility is constrained by the computationally intensive search required to find an optimally compressed adversarial prompt, especially for very long target sequences. Crucially, while these state-of-the-art frameworks differ in their approach, they share a common conceptual limitation: they define memorization through the lens of finding a single elicitation path. They are therefore less designed to characterize the broader accessibility and robustness of a memory, properties that indicate how deeply it is ingrained within the model.

% Introduction of our work
\paragraph{Contributions} In this work, we propose that memories within LLMs can be retrieved using various distinct cues, similar to the concept of cued recall in human memory~\citep{tulving1966availability, tulving1968effectiveness}. We present a framework that redefines memorization, not just by the ability to elicit memories, but by the diversity of paths that can lead to such retrieval. Our main argument is that sequences that are deeply memorized are identifiable by their retrieval through numerous unique prefix prompts. This notion of multiple retrieval avenues is akin to adversarial robustness analysis, where the vulnerability of a model is measured by the size of the input region leading to a particular failure mode \citep{jin2019bert}. We suggest that a greater number of unique prefixes indicates a more deeply memorized sequence, possibly due to a higher frequency of training, and thus presents a higher privacy risk by expanding the potential attack surface.

To implement this idea, our framework evaluates memorization using two connected criteria: an internal recall signal and an external measure of adversarial elicitability. Initially, we determine a \textbf{memorization score ($\eta$)} for a specific sequence, utilizing the model's internal token probabilities. This score is not a conclusive result but is used to establish a systematic, data-dependent \textbf{burden of proof ($P$)}, the required number of distinct adversarial prefixes that need to be identified. The final classification of memorization depends on successfully fulfilling this burden through an external search.

We demonstrate the validity of our framework through the feasibility of this external search. Our experiments show a clear distinction: finding adversarial prefixes for memorized sequences is consistently successful, while the success rate for non-memorized sequences is nearly zero across several model sizes and attack methods. Importantly, this finding also provides a practical computational benefit: auditors can use a persistent failure to find prefixes as a reliable early stopping criterion, allowing them to confidently terminate expensive searches and thereby manage computational resources effectively.

\section{Related Work}
\vspace{3.5mm}
LLM memorization has been characterized through varying definitions, each with specific operational criteria and limitations. We review the most relevant to situate our work.

\paragraph{Discoverable memorization}
A direct approach to measuring memorization is {\em discoverable memorization}, which frames memorization as a natural prefix-completion task \citep{CarliniIJLTZ23}. Given a model $f_\theta$ and a training sequence $s$, $s$ is discoverably memorized if there exists a natural prefix $p$ s.t. $s = [p \mathbin{|} t]$ and greedy decoding from $p$ yields $t$, i.e., $f_\theta(p) = t$. While this method is straightforward and efficient, it presents two significant limitations:\vspace{3mm}
\begin{itemize}

    \item \textbf{Susceptibility to alignment:} As noted by \citet{SchwarzschildFM24}, an LLM fine-tuned to refuse completions of sensitive or known training examples can pass this test, creating an ``illusion of compliance" while the memory may still persist and be accessible via other means.
    
    \item \textbf{Inconsistent correlation with model scale:} The original work suggests a strong positive correlation between model size and discoverable memorization, noting ``a tenfold increase in model size corresponds to an increase in memorization of 19 percentage points". However, this trend may not generalize across datasets and model families. In our preliminary analysis using the Famous Quotes dataset~\citep{SchwarzschildFM24} and the Pythia model suite~\citep{BidermanSABOHKP23}, we observe the opposite: discoverable memorization {\em decreases} with model size, with the largest model (Pythia-12B) exhibiting the lowest memorization rate (Figure~\ref{fig:discoverable_memorization_pythia_quotes}).
    
    This implies that discoverable memorization may not universally reflect a model's underlying tendency to retain training data.
    
\end{itemize}
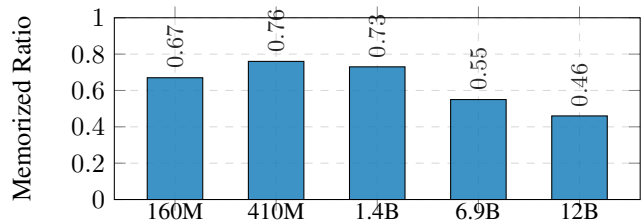
\begin{figure}[t]
\centering
\begin{tikzpicture}
\begin{axis}[
    ybar,
    width=0.5\textwidth,
    height=4cm,
    enlarge x limits=0.15,
    ylabel={Memorized Ratio},
    symbolic x coords={160M, 410M, 1.4B, 6.9B, 12B},
    xtick=data,
    xticklabel style={
        anchor=center,
        font=\small
    },
    nodes near coords, % This command prints the value on top of each bar
    nodes near coords style={
        font=\footnotesize,
        rotate=90, 
        anchor=west,
        /pgf/number format/.cd, % Use PGF's number formatting tools
            fixed,
            precision=3
    },
    ymin=0,
    ymax=1.0, % Set y-axis from 0 to 1 for ratios
    bar width=21pt,
    grid=major,
    grid style={dashed,gray!30}
]

\addplot[fill=blue3f, draw=black, fill opacity=0.85] coordinates {
    (160M, 0.67)
    (410M, 0.76)
    (1.4B, 0.73)
    (6.9B, 0.55)
    (12B, 0.46)
};

\end{axis}
\end{tikzpicture}\vspace{-3mm}
\caption{Discoverable memorization rates for Pythia on Famous Quotes. The ratio reflects quotes elicited via prefix completion. Contrary to expectations, memorization does not scale with model size and declines for largest models. The x-axis signifies the model (Pythia-) size.}\vspace{-3mm}
\label{fig:discoverable_memorization_pythia_quotes}
\end{figure}

\paragraph{Extractable memorization} Another approach to measuring memorization is {\em extractable memorization}, which defines a sample $y$ from the training data as memorized by an LLM $f_\theta$ if an adversary can find any prompt $p$ that elicits $y$ in response, i.e., $f_\theta(p) = y$~\citep{NasrRCHJCICTL25}. While this definition captures adversarial extraction's idea, it suffers from a critical limitation: trivial extraction. This issue arises because the definition can be satisfied by simple instruction-following prompts. For example, given the prompt \textit{``Repeat the following: $y$"}, any compliant model will output $y$, regardless of whether it was truly memorized from the training data. This renders the definition overly inclusive, potentially labeling an entire training set as memorized.

\begin{table*}[t]
\centering
\caption{Comparison of memorization frameworks}
\label{tab:memorization_comparison}\vspace{-3mm}
\scalebox{0.95}{
\begin{tabular}{@{}lcccc@{}}
\toprule
\textbf{Framework} & \textbf{Core Question} & \textbf{Alignment Robust?} & \textbf{Metric} & \textbf{Limitation} \\ \midrule
\textbf{Discoverable} & 
Prefix-completable? & 
No & 
Binary & 
Breaks on aligned models \\

\textbf{Extractable} & 
Elicitable by any prompt? & 
N/A & 
Binary & 
Over-inclusive; trivial prompts \\

\textbf{Compressible} & 
Shorter prompt works? & 
Yes & 
Ratio (ACR) & 
Costly; threshold tuning \\

\textbf{Multi-prefix (Ours)} & 
\# of distinct prompts? & 
Yes & 
Count & 
Computational cost \\ \bottomrule
\end{tabular}
}\vspace{-5mm}
\end{table*}

\paragraph{Compressible memorization} \citet{SchwarzschildFM24} introduced {\em compressible memorization}, framing memorization as a form of efficient compression. Formally, a target sequence $y$ is considered memorized by LLM $f_\theta$ if its adversarial compression ratio (ACR) exceeds a given threshold $\tau$. The ACR is defined as the ratio of the target's length to the length of the shortest possible prompt $x^*$ that elicits it:
\begin{eqnarray}
\nonumber\label{eq:acr} % A label for referencing
\text{ACR}(f
_\theta, y) = \frac{|y|}{|x^*|}; \quad x^* = \arg\min_{x} |x| \text{ s.t. } f_\theta(x) = y.
\end{eqnarray}
A sequence $y$ is then defined as $\tau$-\textit{memorized} if $\text{ACR}(f_\theta, y) > \tau$. The original work suggests a practical threshold of $\tau = 1$. While this definition is more resilient to surface-level alignment, it introduces practical challenges:
\begin{itemize}
    \item \textbf{Computational cost:} A limitation of this framework lies in its high and unpredictable computational cost, stemming from the unknown optimal prompt length $|x^*|$ for a given sequence. Identifying this length requires an iterative meta-search, e.g., the MINIPROMPT algorithm, which repeatedly invokes the costly greedy coordinate gradient (GCG) method~\citep{ZouWKF23} across multiple prompt lengths to locate the success–failure boundary. As a result, the overall runtime reflects not a single optimization, but a sequence of them, governed by the target sequence's unknown characteristics. This cost variability poses a major obstacle to adoption in large-scale auditing where resource predictability is critical.
    
\item \textbf{Arbitrary $\tau$ threshold}: The choice of the threshold $\tau$ is a critical yet subjective aspect of this framework. Both the default $\tau=1$ and the proposed model-agnostic baselines, such as GZIP and SMAZ, are unable to reliably disentangle true memorization from strong generalization. This is because a model-agnostic benchmark, which assesses syntactic patterns, cannot account for the powerful, semantic compression capabilities inherent to a specific LLM's architecture and training. Consequently, a robust metric requires a model-dependent threshold normalized against the model's own baseline performance.
\end{itemize}

\paragraph{Membership inference attacks on foundation models} Our work also relates to the broader field of privacy attacks, particularly membership inference attacks (MIAs). The goal of MIA is to determine whether a specific data point was included in a model's training set \citep{Shokri2017, Yeom2018, Salem2019}. With the rise of LLMs, significant research has focused on adapting MIAs to this new paradigm \citep{MozaffariM2024, fu2024membership, Xie2024}. However, the validity of many recent attacks has been questioned. Several studies argue that their reported success often stems not from detecting genuine membership signals, but from methodological flaws, such as distributional or temporal shifts between the member and non-member evaluation datasets \citep{Duan2024, Maini2024, Meeus2024}. Reinforcing this critique,~\citet{Das2024} demonstrated that even a ``blind" baseline model can outperform these sophisticated attacks, suggesting they primarily exploit dataset artifacts. This recent re-evaluation of MIAs underscores the need for more robust and practical methods for assessing data usage in foundation models, a gap our work aims to address.

\paragraph{Comparative summary}
To synthesize the preceding discussion and clearly position our contribution, we present a comparative summary of memorization frameworks in Table~\ref{tab:memorization_comparison}. While prior work has focused on establishing the possibility (discoverable/extractable) or efficiency (compressible) of eliciting a sequence, our framework introduces a new axis of measurement: the diversity of elicitation paths.

This comparison shows that our multi-prefix framework is not just an alternative but a conceptually distinct approach. By measuring the number of adversarial prompts that elicit a target sequence, we provide a finer-grained assessment of memorization depth. While our method shares the computational demands typical of adversarial search-based techniques, it introduces a principled mechanism for cost control. Notably, our experiments show that adversarial success on non-memorized content is near zero, offering a clear empirical stopping criterion. Consistent failure to elicit a sequence allows auditors to terminate the search early, significantly reducing overhead. This built-in failure signal sets our framework apart from others lacking such a safeguard, making it a more robust and practical tool for auditing LLMs.

\section{Methodology}
\label{sec:methodology}

To quantify the memorization of a particular sequence \(s\) within an LLM \(f_\theta\), our method is grounded in the notion that genuinely memorized sequences by a model can be retrieved through several different input prompts, or prefixes. We assert that this ability to elicit a sequence from multiple prefixes indicates a more robust embedding of the sequence in the model's parameters compared to sequences that are not memorized or are merely generated compositionally. We begin by outlining the conceptual foundation, then present the formal criteria for memorization, and finally detail the computation of its components.

\subsection{Conceptual Framework and Formal Definition}
\label{subsec:conceptual_and_formal}

Our central hypothesis is that the extent to which a sequence \(s\) is memorized by \(f_\theta\) is related to the number of distinct prefixes that can cause \(f_\theta\) to reproduce \(s\) exactly. Although common phrases or highly probable continuations might also be generated from several starting points, we argue that sequences memorized verbatim, especially those less likely under normal compositional generation, will show a significantly greater density of such eliciting prefixes relative to their inherent complexity. Conversely, finding multiple distinct prefixes for specific, non-memorized sequences is expected to be considerably more challenging.

A sequence \(s\) is memorized by \(f_\theta\) if two conditions are fulfilled. First, a \textit{prefix elicitation threshold} must be met: an evaluator must be capable of identifying a set \(\mathcal{X}\) containing at least \(P^s_{f_\theta}\) distinct prefixes, where for each prefix \(p \in \mathcal{X}\), the model's output \(f_\theta(p)\) precisely matches the sequence \(s\). To exclude trivial variations and ensure each prefix's uniqueness, we require that any pair of prefixes for the same target surpass a minimum threshold of cosine distance between their sentence embeddings. The number of distinct prefixes required, denoted as \(P^s_{f_\theta}\), is determined by the sequence's characteristics and its internal memorization signal:
\begin{align}
    P^s_{f_\theta} = \lceil \eta^s_{f_\theta} \times |s| \rceil
    \label{eq:threshold_P}, 
\end{align}
where \(|s|\) represents the length of the sequence \(s\), and \(\eta^s_{f_\theta}\) is the proposed \textit{memorization score}, which measures the intrinsic evidence of memorization (detailed below).

The second condition requires a non-zero internal signal: the calculated memorization score \(\eta^s_{f_\theta}\) must be positive. In our experiments, this corresponds to setting the minimum threshold \(\eta_{min}\) to zero, so the condition becomes $\eta^s_{f_\theta} > 0$.

This requirement ensures that a sequence can only be considered for memorization if there is at least a minimal detectable internal memorization signal. If a sequence has no internal memorization signal (i.e., \(\eta^s_{f_\theta} = 0\)), it will not satisfy this condition and is therefore not considered memorized. In this scenario, \(P^s_{f_\theta}\) would also be zero according to Equation~\ref{eq:threshold_P}. This approach means that the main process for determining the necessary evidence level (i.e., the number of prefixes \(P^s_{f_\theta}\)) depends on the strength of a positive \(\eta^s_{f_\theta}\) and the sequence length \(|s|\).

The prefix elicitation threshold formula \(P^s_{f_\theta}\) is a structured heuristic developed to balance two key elements. It scales with the inherent memorization evidence denoted by \(\eta^s_{f_\theta}\), implying that a higher memorization score (stronger internal recall signals) justifiably requires more external validation through a greater number of unique eliciting prefixes. Additionally, it scales linearly with the complexity of the sequence, represented by its length \(|s|\). This aligns with the intuition that precisely replicating longer sequences by chance is less likely, necessitating more substantial evidence (more prefixes) to claim memorization for longer sequences, even if they possess the same  memorization score \(\eta^s_{f_\theta}\) as a shorter sequence. Though there could be other formulations for \(P^s_{f_\theta}\), this linear scaling provides a straightforward, interpretable method that aligns with these insights, and its success in distinguishing memorized sequences is empirically supported in our experimental section.

It should be noted that this method defines the memorization criterion based on the presence of \(P^s_{f_\theta}\) distinct prefixes. The practical challenge of finding these prefixes is a separate issue that typically involves adversarial search or extraction techniques (e.g., beam search, gradient-based prompt optimization). Our framework offers a principled target number of prefixes, \(P^s_{f_\theta}\), to guide such search efforts, and the specific methods used to find these prefixes in our assessments are detailed in the experiment section.

\subsection{Calculating the Memorization Score}
\label{subsec:memorization_score_calculation} % Changed label to be more unique

We define the memorization score \(\eta^s_{f_\theta}\) to quantify the degree to which the model \(f_\theta\) has internalized the target sequence \(s\). This score integrates two distinct aspects of model behavior, assessed by prompting the model with increasingly longer initial segments of \(s\): a \textit{verbatim recall score} ($r_i$) and a \textit{positional similarity score} ($sim_i$). We derive the overall score \(\eta^s_{f_\theta}\) by aggregating these components across all prefix lengths $i \in [1, |s| - 1]$.

The verbatim recall score $r_i$ aims to measure the model's confidence and inherent tendency to regenerate the remainder of sequence \(s\) (i.e., \(s_{i+1:|s|}\)) verbatim from its initial \(i\) tokens (\(s_{1:i}\)). It is calculated as:
\begin{align}
    \nonumber r_i = \left(\frac{|s| - i}{i}\right) \times p_{avg}(i) \quad \text{for } 1 \le i < |s|.
    \label{eq:recall_score}
\end{align}
The scaling factor \( \frac{|s| - i}{i} \) is chosen to prioritizes recall based on minimal context (small \(i\)). This is because successfully outputting a sequence from a very short prefix is considered stronger evidence of rote memorization than recalling it from a long prefix where the continuation is often more constrained and thus easier to predict. The diminishing weight for larger \(i\) reflects this. While this scaling can be aggressive for very small \(i\), its impact is moderated by \(p_{avg}(i)\) (which is often low for small \(i\)), the subsequent multiplication by the similarity score \(sim_i\), and the final averaging step in the calculation of \(\eta^s_{f_\theta}\). The term \(p_{avg}(i)\) represents the model's average confidence in auto-regressively generating the true continuation \(s_{i+1:|s|}\), calculated as the geometric mean of the conditional probabilities of each correct token in the remainder, given all preceding true tokens:
\begin{eqnarray}
    \nonumber p_{avg}(i) = \exp\left(\frac{1}{|s|-i}\sum_{t = i}^{|s|-1}\log(p(s_{t+1} | s_{1:t}, f_\theta) + \epsilon)\right).
    \label{eq:p_avg}
\end{eqnarray}
Here, \(p(s_{t+1} | s_{1:t}, f_\theta)\) is the model's probability for the true token \(s_{t+1}\) given the preceding context \(s_{1:t}\). A small smoothing value \(\epsilon\) (e.g., \(1 \times 10^{-9}\)) is added before taking the logarithm to handle potential zero probabilities and for stability.

The positional similarity score $sim_i$ measures the model's actual generated output compared to the ground truth sequence \(s\) when prompted with \(s_{1:i}\). Let \(s'_{i+1:|s|}\) be the sequence generated by \(f_\theta\) when prompted with \(s_{1:i}\), then \(sim_i\) is the fraction of tokens in the model's generated continuation that match ground truth sequence \(s_{i+1:|s|}\) at the corresponding positions:
\begin{eqnarray}
    \nonumber sim_i = \frac{1}{|s|-i} \sum_{t=i}^{|s|-1} \mathbb{I}(s'_{t+1} = s_{t+1}) \quad \text{for } 1 \leq i < |s|, \label{eq:similarity_revised}
\end{eqnarray}
where \(s'_{t+1}\) is the generated token at position \(t+1\), \(s_{t+1}\) is the ground truth token, and \(\mathbb{I}(\cdot)\) is the indicator function. This provides a direct measure of the positional accuracy of the model's continuation.

We compute the memorization score \(\eta^s_{f_\theta}\) as the average of the product of these verbatim recall scores (\(r_i\)) and positional similarity scores (\(sim_i\)) across all non-trivial prefix lengths of \(s\) defined as $\eta^s_{f_\theta}= \alpha_1 \sum_{i=1}^{|s|-1} r_i \times sim_i$, or:
\begin{eqnarray}
    \nonumber\eta^s_{f_\theta} = \alpha_1 \sum_{i=1}^{|s|-1} \frac{|s|-i}{i} p_{avg}(i) \times \left( \alpha_i \sum_{t=i}^{|s|-1} \mathbb{I}(s'_{t+1} = s_{t+1}) \right), 
    \label{eq:eta_final_revised}
\end{eqnarray}
where $\alpha_1=\frac{1}{|s|-1}$ and $\alpha_i=\frac{1}{|s|-i}$. 
The averaging over \(i=1\) to \(|s|-1\) serves as a normalization step. It is important to note that due to the scaling factor in \(r_i\), \(\eta^s_{f_\theta}\) is not strictly guaranteed to be within the [0, 1] range. However, its primary purpose is to provide the necessary relative measure of memorization strength. This score is then used to calculate the prefix elicitation threshold \(P^s_{f_\theta}\) as defined in (\ref{eq:threshold_P}). A higher \(\eta^s_{f_\theta}\) value reflects stronger combined evidence from the model's internal confidence (via \(p_{avg}(i)\)) and its external generation accuracy (via \(sim_i\)), with a particular emphasis on recall from shorter, less informative contexts.

\section{Experimental Evaluation}
\label{sec:experiment}
We validate our multi-prefix framework for quantifying sequence memorization and evaluate its effectiveness in distinguishing memorized from non-memorized content. Then, we use it to analyze the impact of instruction fine-tuning, memorization trends across model scales, and patterns of partial memorization within sequences. Finally, we provide an analysis of the computational cost of our framework.

\paragraph{Models and datasets} We use the Pythia model suite~\citep{BidermanSABOHKP23} for our primary analysis because it is open-source and was trained on the publicly documented Pile dataset~\citep{GaoBBGHFPHTNPL21}. We evaluate models across a range of sizes: 160M, 410M, 1.4B, 6.9B and 12B. Our evaluation focuses on two datasets sourced from \cite{SchwarzschildFM24}, both subsets of The Pile. The ``Famous Quotes'' dataset contains 95 quotations and serves as our primary testbed for memorization, while the Wikipedia dataset was used to test memorization on long-form prose.
To assess memorization in instruction-tuned models, we evaluate the Llama-2-7B-chat~\citep{TouvronLK23}, Mistral-7B-Instruct-v0.3~\citep{JiangSMB23}, and Qwen3-14B-Chat~\citep{AnABB25} models against their respective base models: Llama-2-7B, Mistral-7B-v0.3, and Qwen3-14B-Base.
\vspace{-2mm}
\paragraph{Prefix search strategy}\label{p: search_strategy} We employ the GCG algorithm~\citep{ZouWKF23} to search for adversarial prefixes. For each target sequence $s$, we allocate GCG a budget of $\max(10, 2 \cdot P^s_{f_\theta})$ runs, with each run initialized using a different random seed to encourage diverse prefix discovery. The search for $s$ terminates once the required number of distinct prefixes $P^s_{f_\theta}$ is found or when the budget is exhausted. A sequence is ultimately classified as memorized by a model if GCG successfully identifies $P^s_{f_\theta}$ distinct prefixes that cause the model to generate the sequence verbatim.

\subsection{Validation of Multi-Prefix Memorization}
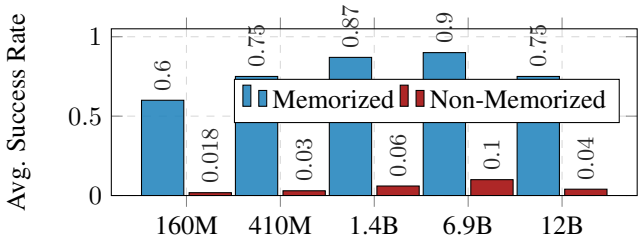
\begin{figure}[t]
\centering
\begin{tikzpicture}
\begin{axis}[
    ybar,
    width=0.5\textwidth,
    height=3.8cm,
    enlarge x limits=0.2,
    legend style={at={(0.6, 0.7)}, anchor=north, legend columns=-1},
    ylabel={Avg. Success Rate},
    symbolic x coords={160M, 410M, 1.4B, 6.9B, 12B},
    xtick=data,
    % Labels will now be horizontal by default
    nodes near coords,
    nodes near coords style={
        /pgf/number format/fixed,
        /pgf/number format/precision=3,
        font=\footnotesize,
        rotate=90, anchor=west
    },
    ymin=0,
    ymax=1.05,
    bar width=16pt,
    grid=major,
    grid style={dashed,gray!30}
]

% --- Data for MEMORIZED sequences ---
\addplot[fill=blue3f,
        draw=black,
        fill opacity=0.85] coordinates {
    (160M, 0.60)
    (410M, 0.75)
    (1.4B, 0.87)
    (6.9B, 0.90)
    (12B, 0.75)
};

% --- Data for NON-MEMORIZED sequences ---
\addplot[fill=red3f,
        draw=black,
        fill opacity=0.85] coordinates {
    (160M, 0.018)
    (410M, 0.03)
    (1.4B, 0.06)
    (6.9B, 0.10)
    (12B, 0.04)
};

\legend{Memorized, Non-Memorized}
\end{axis}
\end{tikzpicture}\vspace{-2mm}
\caption{GCG success rates by model size and sequence type. Memorized sequences consistently yield higher success rates, validating the classification method and revealing the attack's dependence on prior memorization.}
\label{fig:combined_memorization_success_rate}\vspace{-3mm}
\end{figure}

To assess the core concept of our approach, that memorized sequences are more susceptible to elicitation, we analyze the attack success rate (ASR) of our prefix discovery method. Figure~\ref{fig:combined_memorization_success_rate} visualizes this by plotting the ASR for both ``memorized" and ``non-memorized" sequences side-by-side for each model in the Pythia suite. The results reveal a strong disparity. For sequences classified as memorized, the ASR is consistently high and scales with model size, rising from 0.60 for Pythia-160M to 0.90 for Pythia-6.9B. In contrast, the success rates for non-memorized sequences are dramatically lower across the board. For the Pythia-160M model, this rate is a mere 0.018, implying an average of 56 GCG runs would be needed to find a single valid prefix. This side-by-side comparison provides compelling evidence that our methodology effectively distinguishes between sequences that are deeply embedded and easily elicitable versus those that are not.

\begin{figure}[t] % It's good practice to use [h!] to encourage placement here
    \centering
        \begin{tikzpicture}
            \begin{axis}[
                ybar,
            width=0.5\textwidth,
            height=3.8cm,
                bar width=22pt, % <-- Made bar width smaller to look better
                enlarge x limits=0.5,
                title style={font=\small}, % <-- Removed bold, used small for consistency
                ylabel={Mem. Rate (\%)},
                xlabel={}, % <-- ADDED empty label to balance vertical space
                symbolic x coords={Original, Paraphrased},
                xtick=data,
                nodes near coords,
                nodes near coords align={vertical},
                every node near coord/.append style={font=\footnotesize}, % <-- Added for consistency
                ymin=0,
                ymax=102,
                ymajorgrids=true,
                grid=major,
                grid style={dashed,gray!30},
                tick label style={font=\footnotesize}, % <-- Added for consistency
                label style={font=\footnotesize}
            ]
            \addplot[fill=blue3f, draw=black, fill opacity=0.85] 
                coordinates {(Original, 84.2) (Paraphrased, 3.2)};
            \end{axis}
        \end{tikzpicture}

    \begin{tikzpicture}
    \begin{axis}[
        % --- Main plot styling ---
            width=0.5\textwidth,
            height=3.8cm,
        ylabel={\# Quotes},
        xlabel={Recall Rate (\%)},
        % --- Bar chart specific styling ---
        ybar,
        bar width=10pt,
        enlarge x limits=0.05,
        % --- X-axis styling for categories ---
        symbolic x coords={0, 10, 11, 12, 14, 16, 20, 30, 33, 37, 42, 50, 83, 100},
        xtick=data,
        % xticklabel style={rotate=45, anchor=east}, % <-- ADDED to prevent label overlap
        % --- General styling ---
        ymin=0,
        ymax=30,
        ymajorgrids=true,
        nodes near coords,
        every node near coord/.append style={font=\footnotesize},
        tick label style={font=\small},
        label style={font=\small},
        title style={font=\small},
        grid=major,
        grid style={dashed,gray!30}
    ]
    % --- Bar Chart Data ---
    \addplot[fill=red3f, draw=black, fill opacity=0.85] table[x=rate, y=count] {
        rate count
        0 1
        10 10
        11 22
        12 14
        14 12
        16 6
        20 2
        30 5
        33 7
        37 7
        42 6
        50 1
        83 1
        100 1
    };
    \end{axis}
    \end{tikzpicture}
\caption{Memorization behavior in Pythia-6.9B for original Famous Quotes and paraphrased quotes. \textbf{Top:} \textit{Memorization rates for original vs.\ paraphrased quotes. Minimal paraphrasing drops memorization from 84\% to 3.2\%, showing the method detects verbatim memorization.} \textbf{Bottom:} \textit{Distribution of recall rates. Adversarial attacks on paraphrased quotes often trigger the model to recall and output the original memorized quote.}}
\label{fig:full_results}
\end{figure}
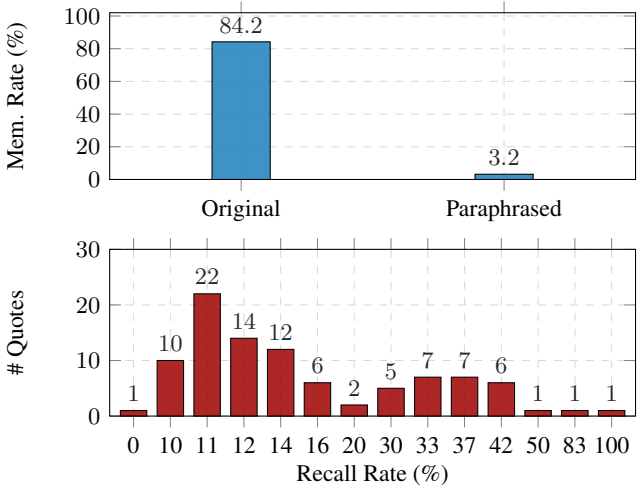

\paragraph{Memorization of paraphrased Famous Quotes}
To determine if our definition can separate between exact and conceptual memorization, we tested the Pythia-6.9B model on both {\em Famous Quotes} and a {\em paraphrased version}. For the paraphrased set, we used the Gemini 2.5 Pro API~\citep{Gemini} to alter a maximum of two words per quote while preserving its meaning. The results are shown in Figure~\ref{fig:full_results} (top part). The model, which memorized approximately $84\%$ of the original quotes, memorized only 3 of the 95 paraphrased versions. This disparity shows that our method detects exact, not conceptual, memorization.
\pgfplotstableread[col sep=space]{
x_start y_count
0.0 1
7.7 58
15.4 8
23.1 5
30.8 14
38.5 6
46.2 1
53.8 0
61.5 0
69.2 0
76.9 1
84.6 0
92.3 1
}\myhistdata

We observed a further, more interesting phenomenon during GCG attacks. In particular, the adversarial prompts designed for the paraphrased quotes consistently caused the model to output the \textit{original} famous quotes instead. In fact, as shown in Figure~\ref{fig:full_results} (bottom part), for 94 out of the 95 paraphrased quotes, the algorithm successfully found an adversarial prompt that elicited the original quote. This high success rate (as shown in Figure~\ref{fig:full_results}; bottom part) is reflected in the overall metrics, where the ratio of found original quotes to GCG attempts was nearly 20\%. This serves as a strong cue that the model's memory of these phrases is tied to the exact original text.

\paragraph{Memorization of random sequences}
Since our method is shown to be able to distinguish between exact and conceptual memorization, we want to ensure that our method specifically identifies memorized content rather than producing adversarial prompts for arbitrary sequences. To this end, we conducted a control experiment. We generated a dataset of 1,000 random token sequences by uniformly sampling with replacement from the model's vocabulary, with lengths ranging from 5 to 15 tokens. These sequences decode to semantically meaningless gibberish and serve as non-memorized controls. Across all evaluated model sizes, our algorithm failed to find any prefix that elicited these sequences, confirming that our measurements reflect true memorization rather than artifacts of the attack method.

\paragraph{Adversarial prefix finding}
Besides GCG, we leverage the probe sampling (PS) method \citep{ZhaoZCLKGS24} to enhance the robustness of the GCG algorithm. The original paper reports that PS increases the attack success rate (ASR), partly due to the randomness it introduces. While PS is observed to improve the speed of the GCG algorithm, we found that for our goal of discovering a diverse set of prefixes, the GCG+PS combination is a more conservative method that finds fewer unique prefixes than GCG alone. For example, on the Pythia-6.9B model, 21 sequences were classified as non-memorized when tested with GCG+PS, compared to only 15 when tested with GCG. Similarly, on the 12B model, 8 out of 95 sequences that were classified as memorized under our definition using the GCG, were subsequently classified as non-memorized when tested with GCG+PS. Despite this, the average attack success rate for sequences classified as memorized remains high, at 0.76 for the 6.9B model and 0.73 for the 12B model. For those classified as non-memorized, the average success rates are 0.06 and 0.08. 

\paragraph{Statistical validation} To statistically validate our multi-prefix memorization framework, we tested its core hypothesis ($H_a$): sequences classified as ``memorized" are significantly more susceptible to adversarial elicitation than those classified as ``non-memorized.", while the null hypothesis $H_0$ states that there is no significant difference in GCG attack success rates between the two classes. We conducted a one-sided Mann-Whitney U test on the GCG attack success rates for the two classes, using the Famous Quotes dataset. By evaluating on models of different scales, we demonstrate the framework's robustness and generalizability.

First, on the large Pythia-6.9B model, our framework achieved a near-perfect separation between the two classes. Sequences labeled "memorized" had a median GCG success rate of 1.0, while "non-memorized" sequences had a median of 0.0. This visual distinction was confirmed to be statistically significant $(U = 1189.0, p < 1e-10)$. Furthermore, the effect size was exceptionally large (Rank-biserial correlation $r \approx 0.99$), indicating that the classification provided a practically complete separation between elicitable and non-elicitable content.

Next, to test the framework's robustness on a less capable model, we repeated the analysis on the small Pythia-160M model. Even here, where memorization is far less prevalent, our method proved effective. The ``memorized" group exhibited a median success rate of 0.5, compared to 0.0 for the ``non-memorized" group. Despite the weaker memorization signals in the model, the distinction between the groups remained highly significant $(U = 1457.5, p < 1e-10)$, again with a large effect size $(r = 0.89)$. The consistent statistical significance and large effect sizes across models of different scales provide strong empirical evidence for our framework. It confirms that our methodology robustly identifies a meaningful and practical distinction between memorized and non-memorized sequences.

\subsection{Adversarial Prefixes Characterization}
\paragraph{Semantic diversity of adversarial prefixes}
To characterize the mechanism of model memorization, we first evaluate the semantic diversity of adversarial prefixes that elicit the same target sequence. We hypothesize two potential outcomes: the prefixes could be semantically clustered (low pairwise distance), suggesting the optimization converges on a narrow semantic region, or they could be semantically scattered (high pairwise distance). The latter case would imply a more arbitrary, lookup-table-like behavior, where disparate inputs are mapped to an identical, verbatim output.

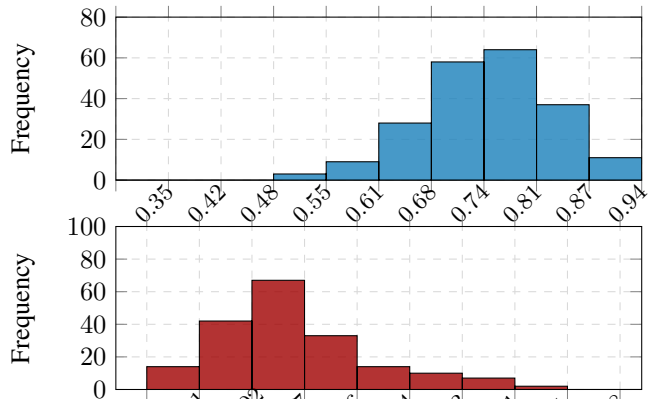
\begin{figure}[h]
    \centering

    % First histogram (top)
    \begin{tikzpicture}
        \begin{axis}[
            ylabel={Frequency},
            width=0.5\textwidth,
            height=3.75cm,
            ybar interval,
            xmin=0.35, xmax=1,
            ymin=0,ymax=80,
            xticklabel style={/pgf/number format/fixed, rotate=45, anchor=north, font=\small},
            grid=major,
            grid style={dashed,gray!30},
        ]
            \addplot[
                fill=blue3f,
                draw=black,
                fill opacity=0.8
            ] coordinates {
                (0.3500, 0) (0.4150, 0) (0.4800, 0) (0.5450, 3) (0.6100, 9)
                (0.6750, 28) (0.7400, 58) (0.8050, 64) (0.8700, 37) (0.9350, 11) (1.0000, 0)
            };
        \end{axis}
    \end{tikzpicture}

    \vspace{-0.4cm} % Add vertical space between the plots

    % Second histogram (bottom)
    \begin{tikzpicture}
        \begin{axis}[
            ylabel={Frequency},
            width=0.5\textwidth,
            height=3.75cm,
            ybar interval,
            xmin=-0.15, xmax=0.7,
            ymin=0, ymax=100,
            xticklabel style={/pgf/number format/fixed, rotate=45, anchor=north, font=\small},
            grid=major,
            grid style={dashed,gray!30},
        ]
            \addplot[
                fill=red3f,
                draw=black,
                fill opacity=0.8
            ] coordinates {
                (-0.1000, 14) (-0.0150, 42) (0.0700, 67) (0.1550, 33)
                (0.2400, 14) (0.3250, 10) (0.4100, 7) (0.4950, 2)
                (0.5800, 0) (0.6650, 0) (0.7500, 0)
            };
        \end{axis}
    \end{tikzpicture}\vspace{-6mm}
    \caption{Semantic similarity statistics for adversarial prefix generation targeting famous quotes. \textbf{Top:} \textit{Distribution of cosine distances between adversarial prefixes of each target (Famous Quotes)}. \textbf{Bottom:} \textit{Distribution of cosine similarities between original quotes and their adversarial prefixes}.}\label{fig:similarity_histogram_target_prefix}\label{fig:distance_histogram_clean}\vspace{-3mm}
\end{figure}

Our analysis focuses on adversarial prefixes optimized for the Pythia-6.9B model to elicit targets from the Famous Quotes dataset. We compute the pairwise cosine distance between the all-MiniLM-L6-v2 embeddings~\citep{Reimers2019SBERT, Wang20minilm} of all successful prefixes for each target and present the aggregated distribution in Figure \ref{fig:distance_histogram_clean}. The distribution is clearly skewed towards higher values, with a significant proportion of prefix pairs exhibiting a cosine distance greater than 0.5. This high degree of semantic diversity suggests that the model has not learned a robust, coherent pathway to the target sequence. This finding strongly supports the memorization hypothesis, indicating a lookup-like retrieval mechanism where multiple, disparate inputs are mapped to the same stored data point, rather than guiding a semantically-grounded generation.

\paragraph{Connection between prefixes and target sequences}
To investigate the operational mechanism by which adversarial prefixes elicit specific target sequences, we computed the cosine similarity between entries from the Famous Quotes dataset and their corresponding adversarial prefixes which were uniquely optimized for the Pythia-6.9B model by GCG. This analysis sought to determine whether the prefixes function as meaningful contextual prompts or as arbitrary token sequences that exploit learned statistical patterns. A high degree of semantic correspondence would indicate a mechanism of contextual guidance, whereas a lack of correspondence would imply a non-semantic retrieval process.

The results reveal a lack of semantic correlation between the prefixes and their targets. The frequency distribution of similarity scores, presented in Figure~\ref{fig:similarity_histogram_target_prefix}, is unimodal and heavily right-skewed, indicating that the vast majority of prefixes are semantically dissimilar to the content they produce. The distribution exhibits a pronounced mode within the low-similarity interval of [0.070, 0.155). Furthermore, over 82\% of all analyzed prefixes yield a cosine similarity score below 0.24. This finding strongly points to verbatim memorization. The absence of a semantic link rules out contextual guidance, suggesting the prefix functions not as a prompt, but as a key in a system analogous to a lookup table. In this mechanism, a specific, arbitrary input retrieves a corresponding value, which in this case is a sequence memorized verbatim from the model's training data.

\subsection{Memorization Across Model Sizes}
\begin{figure}[h]
\centering
\begin{tikzpicture}
\begin{axis}[
    ybar, % Use ybar style for the plot
    width=0.5\textwidth,
    height=3.75cm,
    enlarge x limits=0.15,
    ylabel={Mem. Ratio (\%)},
    symbolic x coords={160M, 410M, 1.4B, 6.9B, 12B},
    xtick=data,
    ymin=0,
    ymax=100,
    % Add nodes near coords to display the value
    nodes near coords,
    % Align the text vertically above the bar for a clean look
    nodes near coords align={vertical},
    bar width=20pt,
    grid=major,
    grid style={dashed,gray!30}
]
% --- The single plot for hollow bars with values on top ---
% 'fill=none' makes the bars hollow.
% 'draw=black' makes the outline black.
% 'nodes near coords' (defined above) will automatically apply to this plot.
\addplot[fill=blue3f, draw=black, fill opacity=0.8] coordinates {
    (160M, 20)
    (410M, 40)
    (1.4B, 71)
    (6.9B, 84)
    (12B, 83)
};
\end{axis}
\end{tikzpicture}\vspace{-2mm}
\caption{Proportion of sequences from the Famous Quotes dataset classified as memorized by each Pythia model. The results demonstrate a clear trend of increased memorization with model size. The x-axis signifies the model size.}
\label{fig:memorization_scaling}\vspace{-4mm}
\end{figure}
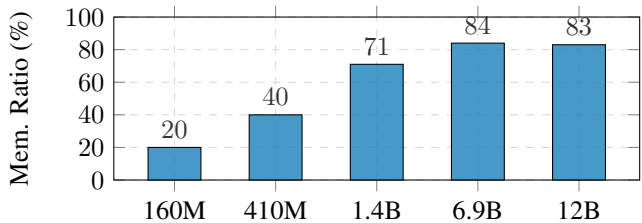

\paragraph{Famous Quotes} We investigated how memorization scales with model size by applying our multi-prefix criterion to the Famous Quotes dataset across the Pythia model suite. As shown in Figure~\ref{fig:memorization_scaling}, the results reveal a strong positive trend: larger models consistently memorize a higher proportion of quotes. Interestingly, this trend appears to plateau at the largest scales, with the 12B model showing a memorization ratio of 83\%. This finding aligns with prior work linking increased model capacities to enhanced memorization~\citep{CarliniIJLTZ23, SchwarzschildFM24}. It also indicates a possible saturation point for this particular kind of verbatim knowledge due to the generalizable abilities of larger models.

\paragraph{Wikipedia} Our Wikipedia experiments reveal a more complex relationship between scale and memorization. While the memorized ratio for the 1.4B model ($\approx$5\%) aligns with the findings of \citep{SchwarzschildFM24} (the only size they reported), we observe a distinct decrease for larger models, with the 6.9B and 12B variants memorizing only around 2\% of sequences. We hypothesize that this divergence is attributable to the nature of the data itself. Unlike the short, canonical strings of famous quotes, Wikipedia text consists of continuous, long-form prose. For such data, larger models may leverage their enhanced capacity for contextual understanding to generalize information rather than resorting to rote memorization of the exact phrasing.

\subsection{Analysis of Partial Sequence Memorization}
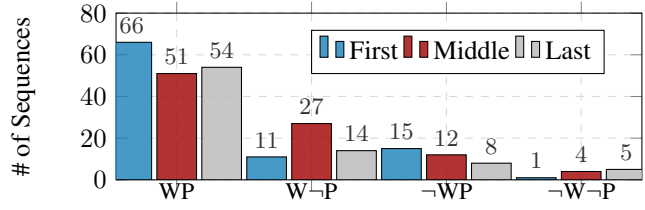
\begin{figure}[h]
\centering
\begin{tikzpicture}
\begin{axis}[
    width=0.5\textwidth,   % Increased width to accommodate grouped bars
    height=3.8cm,
    ybar,                  % Specifies a vertical bar chart
    bar width=15pt,        % Adjust bar width for a good fit
    enlarge x limits=0.15, % Padding on the sides
    ylabel={\# of Sequences},
    symbolic x coords={Both, WholeNotPart, PartNotWhole, Neither}, % Categories on x-axis
    xtick=data,
    xticklabels={
        {\small WP},
        {\small W$\neg$P},
        {\small $\neg$WP},
        {\small $\neg$W$\neg$P}
        % {\small Whole \& Part Mem.},
        % {\small Whole only, Part Not Mem.},
        % {\small Part only, Whole Not Mem.},
        % {\small Neither Mem.}
    },
    xticklabel style={
        %rotate=30,         % Rotated labels to prevent overlap
        anchor=center
        %,font=\small
    },
    nodes near coords,     % Display values on top of bars
    nodes near coords style={font=\footnotesize, /pgf/number format/fixed}, % Make the numbers smaller
    ymin=0,ymax=80,
    legend style={
        at={(0.65, 0.9)},  % Position legend at the top-center inside the plot
        anchor=north,      % Anchor the legend's top edge to the point
        legend columns=-1  % Arrange legend items horizontally
    },
    %title={Memorization: Whole Sequence vs. Partial Halves (Pythia-6.9B)}, % A single title for the combined chart
    grid=major,
    grid style={dashed,gray!30}
]

% Plot for the First Half
\addplot[fill=blue3f, draw=black, fill opacity=0.8] coordinates {(Both,66) (WholeNotPart,11) (PartNotWhole,15) (Neither,1)};

% Plot for the Middle Half
\addplot[fill=red3f, draw=black, fill opacity=0.8] coordinates {(Both,51) (WholeNotPart,27) (PartNotWhole,12) (Neither,4)};

% Plot for the Last Half
\addplot[fill=gray3f, draw=black, fill opacity=0.8] coordinates {(Both,54) (WholeNotPart,14) (PartNotWhole,8) (Neither,5)};

% Add legend entries to describe each bar group
\legend{First, Middle, Last}

\end{axis}
\end{tikzpicture}\vspace{-2mm}
\caption{Memorization across sequence segments in Pythia-6.9B. Bars show outcomes for whole (W) and partial (P: first, middle, final) subsequences. $\neg$ for non-memorized. Memorization varies notably with segment position.}
\label{fig:partial_memorization_comparison}\vspace{-4mm} % A single, new label for the combined figure
\end{figure}

While existing literature assesses memorization at the full sequence level \citep{CarliniIJLTZ23, NasrRCHJCICTL25, SchwarzschildFM24}, we investigate the relationship between whole-sequence memorization and the memorization of its constituent parts. To this end, for each sequence in the Famous Quotes dataset, we extract and analyze three overlapping subsequences, each comprising 50\% of the original's token length: the initial 50\% of tokens (the first half), a central 50\% segment (the middle half), and the final 50\% of tokens (the final half). For example, consider the quote ``I think, therefore I am.". The first half would be ``I think,", the middle half would be ``think, therefore" and the final half would be ``therefore I am." We applied our memorization analysis to these subsequences using the Pythia-6.9B model. The outcomes, detailed in Figure~\ref{fig:partial_memorization_comparison}, reveal that memorization behavior differs significantly depending on the subsequence's position within the original text.

\if0
Our analysis in Figure~\ref{fig:partial_memorization_comparison} reveals a strong correlation between whole-sequence memorization and the memorization of its constituent parts. When a whole sequence is memorized, its initial and final segments are very likely to be memorized as well, with 66 and 54 instances of joint memorization, respectively. However, the middle half demonstrates a notable fragility. Even when the entire sequence meets our memorization criteria, its central portion is the most likely to be forgotten; this ``Whole only, Part Not Mem." outcome occurred in 27 instances for the middle half, a stark contrast to only 11 instances for the first half and 14 for the last.
\fi

Our analysis in Figure~\ref{fig:partial_memorization_comparison} reveals a strong positional bias. When a full sequence is memorized, its initial and final segments are often co-memorized (66 and 54 instances, respectively). The middle segment, however, is notably fragile and is forgotten in 27 instances, far more than the first (11) or final (14) parts. This primacy effect extends to cases of partial-only recall, where the first half is most frequently memorized alone (15 instances), followed by the middle (12) and final (8) halves. This suggests that while full-sequence memorization often includes its parts, initial segments are most robustly and independently retained, whereas central portions are less consistently memorized, even when the whole is recalled.

\vspace{-1mm}
\subsection{Impact of Instruction Fine-Tuning}
\label{sec:finetuning_across_families}

A central question in model development is how instruction fine-tuning impacts the recall of pre-trained knowledge. While one might assume that alignment for conversational tasks would uniformly suppress the verbatim recall of training data, our findings indicate a more complex and unexpected relationship. We propose that this complexity can be best understood by distinguishing between two separate factors: the influence of the alignment process itself and the distinct effect of the conversational chat template. To investigate this, we compared the memorized ratio between the base and chat-aligned versions of Llama-2-7B, Mistral-7B-v0.3, and Qwen3-14B on the Famous Quotes dataset, using our multi-prefix memorization technique and assessing the chat models both with and without their chat templates.

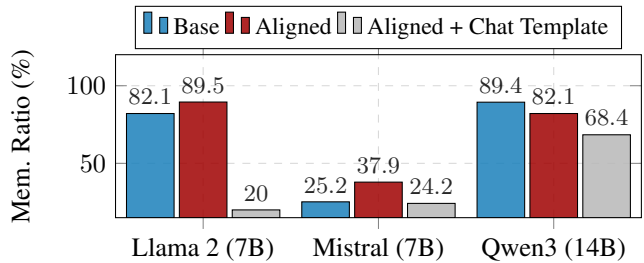
\begin{figure}[h]
\centering
\begin{tikzpicture}
\begin{axis}[
    ybar,
    width=0.5\textwidth,
    height=3.75cm,
    enlarge x limits=0.25,
    legend style={at={(0.5,1.3)}, anchor=north, legend columns=-1, font=\small},
    ylabel={Mem. Ratio (\%)},
    symbolic x coords={Llama 2 (7B), Mistral (7B), Qwen3 (14B)},
    xtick=data,
    nodes near coords,
    nodes near coords align={above},
    nodes near coords style={font=\footnotesize},
    ymin=15,
    ymax=120,
    bar width=18pt,
    grid=major,
    grid style={dashed,gray!30}
]

% --- Data for BASE models ---
\addplot[fill=blue3f, draw=black, fill opacity=0.85] coordinates {
    (Llama 2 (7B), 82.1)
    (Mistral (7B), 25.2)
    (Qwen3 (14B), 89.4)
};

% --- Data for ALIGNED models ---
\addplot[fill=red3f, draw=black, fill opacity=0.85] coordinates {
    (Llama 2 (7B), 89.5)
    (Mistral (7B), 37.9)
    (Qwen3 (14B), 82.1)
};

% --- Data for ALIGNED + CHAT TEMPLATE models ---
\addplot[fill=gray3f, draw=black, fill opacity=0.85] coordinates {
    (Llama 2 (7B), 20.0)
    (Mistral (7B), 24.2)
    (Qwen3 (14B), 68.4)
};

\legend{Base, Aligned, Aligned + Chat Template}
\end{axis}
\end{tikzpicture}\vspace{-2mm}
\caption{Memorization ratios on famous quotes for Llama 2, Mistral, and Qwen3 model families. The effect of alignment is model-dependent: it increases memorization for Llama 2 and Mistral but decreases it for Qwen3. In contrast, applying a chat template consistently and significantly reduces the memorization ratio across all three models.}
\label{fig:llama_qwen_memorization_comparison}\vspace{-4mm}
\end{figure}

\paragraph{Analysis of memorization} Our analysis separates the impact of the alignment process from the application of a chat template, revealing a dynamic in model behavior.
A direct comparison between base and aligned models shows varied effects: for instance, Llama 2's memorization ratio increases from 82.1\% to 89.5\% after alignment, and Mistral's rises from 25.2\% to 37.9\%. Conversely, Qwen3's memorization decreases from a base of 89.4\% to 82.1\% after alignment. However, a consistent pattern emerges when the chat template is applied: it drastically reduces the memorization of the aligned model in all cases, down to 20.0\% for Llama 2, 22.0\% for Mistral, and 68.4\% for Qwen3.

These results highlight the role of the chat template as a defense mechanism. Its structural tokens (e.g., \texttt{[INST]} for Llama 2, \texttt{<|im\_start|>} for Qwen3) invoke conversational protocols that suppress elicitation. When the template is absent, the model reverts to its fundamental text-completion capabilities, exposing a degree of memorization that is otherwise masked. In contrast, the alignment process when evaluated without a chat template has a more complex impact on memorization. For models like Llama 2 and Mistral, it appears to increase the model's tendency to recall stored data, while for Qwen3, it provides a slight reduction, though the memorization level remains high.

This distinction is critical for auditing.
Given that the impact of alignment on memorization varies across models, we suggest that auditors may consider evaluating the aligned model directly, \textit{without} its chat template. This approach can provide a more direct assessment of a model's underlying memorization, offering a clearer signal of its potential to leak training data, independent of the variable effects of its conversational guardrails.

\subsection{Analysis of Computational Cost}

\paragraph{Required number of prefixes}
We constructed a dataset of 900 sequences from the Pile-CC corpus, which is part of the Pile dataset. To ensure a balanced representation of sequence lengths, we employed a stratified sampling strategy: for each word count from 10 to 99 inclusive, we randomly selected 10 unique sequences. This methodology resulted in a dataset where sequence token counts range from 11 to 416, with a mean of 65.9 tokens per sequence.

\begin{figure}[h]
\centering % Center the entire figure environment

    \begin{tikzpicture}
        \begin{axis}[
            width=0.5\textwidth,
            height=3.8cm,
            xlabel={Number of Prefixes},
            ylabel={\# Sequences},
            ybar,
            bar width=13,
            enlarge x limits=0.1,
            ymin=0,
            % ymajorgrids=true,
            grid=major,
            grid style={dashed, color=gray!30},
            xtick=data
        ]
            \addplot+[fill=blue3f, draw=black, fill opacity = 0.85] 
                table[x=num_prefixes, y=frequency] {\dataForFigA};

\end{axis}
    \end{tikzpicture}
    \label{subfig:2a}

    \begin{tikzpicture}
        \begin{axis}[
            width=0.5\textwidth,
            height=3.8cm,
            xlabel={Token Count},
            ylabel={\# Prefixes},
            enlarge x limits=0.1,
            % ymajorgrids=true,
            grid=major,
            grid style={dashed, color=gray!30},
        ]
            % Plot the scatter plot from your data
            \addplot[
                scatter, 
                only marks,
                % mark=*,
                % mark size=1.2pt,
                color=blue3f,
                opacity=0.5, % Opacity helps visualize point density
            ] table[x=token_count, y=num_prefixes] {\dataForFigB};
        \end{axis}
    \end{tikzpicture}
\caption{The number of prefixes required for memorization. \textbf{Top:} \textit{Distribution of required prefixes. Most sequences are memorized with 2--3 prefixes, with a mean of 2.62; the distribution is heavily right-skewed.} \textbf{Bottom:} \textit{Number of required prefixes vs.\ sequence length. There is little to no correlation between token count and prefix requirement; most sequences are memorized with five or fewer prefixes.}}
\label{fig:memorization_analysis}

\end{figure}
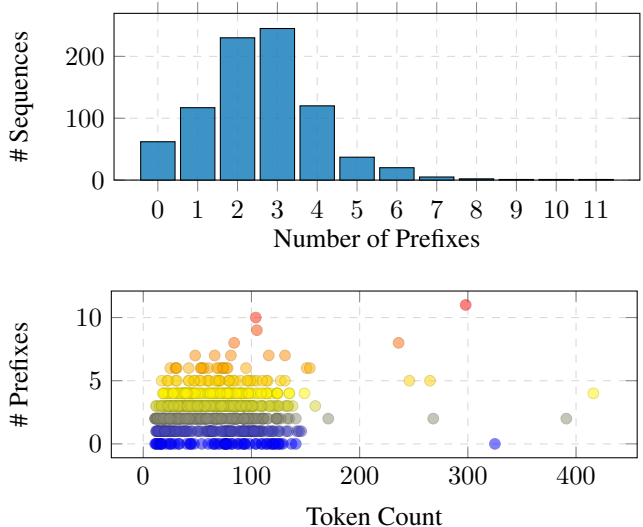

In our memorization analysis of this dataset, we found that on average, only 2.62 prefixes were required to uniquely identify a sequence (See Figure~\ref{fig:memorization_analysis}, top part). More importantly, we observed that the distribution of required prefixes is heavily right-skewed, with the vast majority of sequences needing five or fewer prefixes ($P^s_{f_\theta} \leq 5$) to be counted as memorized (see Figure \ref{fig:memorization_analysis}, bottom part). This empirical finding is the basis of our running budget for each sequence as $max(10, 2 \cdot P^s_{f_\theta})$, to make the computation both efficient and robust. Since our data shows most sequences have $P \leq 5$, the $2\cdot P$ term evaluates to 10 or less for these common cases, causing the budget to default to a stable 10 runs. This approach prevents excessive computational cost for easily memorized sequences. At the same time, the $2 \cdot P$ term serves as a dynamic bound for rare outlier sequences where $P > 5$. This ensures that harder-to-memorize sequences are automatically allocated a proportionally larger budget.

\pgfplotstableread[col sep=space]{
% --- BEGIN DATA FOR FIGURE 2 ---
num_prefixes frequency
0 62
1 117
2 230
3 245
4 120
5 37
6 20
7 5
8 2
9 1
10 1
11 1
}\dataForFigA

% --- Data for Figure 3 (Scatter Plot) ---
% PASTE THE SCATTER PLOT DATA FROM THE PYTHON SCRIPT HERE
\pgfplotstableread[col sep=space]{
% --- BEGIN DATA FOR FIGURE 3 ---
token_count  num_prefixes
   65         2
     101         1
      76         4
      83         1
      12         0
     101         3
     325         0
      53         2
      60         5
     102         3
      17         1
     123         3
      50         2
      28         3
      50         3
      54         6
      30         5
      70         4
      70         3
      43         0
      77         1
     236         8
      54         3
      25         0
     102         3
      55         3
      65         2
      26         3
     123         4
      32         4
      83         1
     123         2
      60         6
      67         1
      69         2
      32         3
      22         5
      45         4
     126         1
     110         3
      12         3
      30         2
      76         1
      95         4
     124         2
      92         2
      27         2
      18         1
      62         2
      35         3
      54         0
      29         2
      71         3
      84         2
      98         3
      36         4
      39         3
      38         3
      93         1
      32         4
      56         3
      20         2
      33         1
      73         6
      72         2
     109         2
      73         3
      79         5
      33         2
     109         2
      75         0
      76         0
      51         3
      72         2
      17         5
      75         2
      25         3
     102         4
      12         3
     119         1
      72         3
      74         1
      26         2
     121         1
      68         3
      14         1
      85         5
      18         2
      79         4
      35         2
      38         1
      67         2
     135         2
     115         4
      19         1
      36         4
      12         2
     111         4
      37         2
      97         4
      80         3
      21         1
      14         2
      76         4
      11         2
      91         4
      22         3
      28         1
      41         2
      17         3
      40         3
      75         0
     106         4
      56         4
      67         3
     125         3
     111         2
      90         5
      54         3
     105         4
      78         1
      35         2
     142         1
      52         3
      37         3
      28         1
      70         6
      13         2
      82         4
      21         2
      34         3
     109         3
      74         3
      49         2
      87         2
      25         6
      92         3
      54         3
      20         4
      93         1
      84         4
      80         3
     118         4
      65         2
      14         1
      30         2
      90         3
      55         3
      86         2
      46         4
      31         4
      19         1
      66         3
      40         2
      90         4
      32         2
     114         4
      83         4
     127         5
      66         7
     146         1
      33         2
      43         3
      48         2
      48         5
      43         3
      78         4
      29         2
     104         2
      87         0
      11         0
      68         2
      44         4
      24         1
      72         3
      39         3
     102         1
      65         3
      57         2
     127         0
      81         4
      81         3
      94         0
      58         2
     102         2
     125         1
     121         3
      56         3
      73         4
      61         4
      74         4
      42         3
      78         3
      92         3
      40         1
      50         2
      93         4
     298        11
     102         5
      53         2
      67         2
      36         4
      65         2
      59         1
      78         4
     141         2
      77         1
      47         2
      51         2
      69         3
      96         1
      53         4
      81         7
      39         4
     117         0
      16         0
      86         2
      60         3
     111         2
      51         3
      83         2
     100         2
     124         4
      81         3
      95         3
      74         3
      68         2
      72         3
      29         0
      80         4
      71         3
     119         0
      27         3
      99         3
      69         0
      56         3
      76         3
      16         2
      17         1
      76         3
      44         3
     131         5
      84         2
      97         5
      21         0
      87         4
      40         0
      54         3
      61         3
      53         3
      80         5
      36         4
      19         2
      31         2
      38         2
      25         3
      80         2
     102         1
      55         2
      50         3
      62         1
      42         3
      14         1
      56         2
      93         2
      93         3
      78         3
      96         1
      34         3
      24         4
      24         3
      14         1
      92         3
      81         1
      86         3
      96         3
     119         3
      25         1
      48         7
      83         3
      20         2
      26         1
     100         1
      57         1
      97         2
      51         3
      27         3
      11         2
      59         4
      17         0
      40         3
      98         1
      62         1
     151         6
      20         3
      85         4
      84         3
      55         4
     108         1
     108         1
      48         3
      63         3
      79         4
      92         2
      74         6
      31         3
      45         1
      51         1
     109         0
      77         1
     114         0
      15         2
      77         3
      69         2
      11         2
      24         3
      50         4
      25         2
      44         3
      70         5
     125         0
      56         2
      34         2
      39         4
      55         3
      82         0
      54         6
      80         3
      90         1
      74         2
      88         0
      49         3
      12         2
      31         2
      66         3
      19         2
      45         2
      96         2
      97         4
      98         2
      91         4
     144         1
     133         2
      19         2
      24         0
     101         3
     104         0
     102         2
      14         2
     116         7
      40         3
      53         3
     127         2
      81         4
      33         0
     121         0
      50         4
      64         3
      94         3
      97         3
      38         3
      93         0
      73         3
     119         1
      98         1
      54         2
      30         6
      43         2
      59         5
      23         4
     129         0
      43         4
      42         5
     110         1
      64         2
      53         2
      63         1
      44         3
      46         3
     100         3
      40         2
      50         3
      49         3
      21         3
      74         4
      79         4
      63         3
      62         3
      84         4
      90         4
      56         3
      13         2
      86         3
      30         2
      70         2
      76         3
      47         3
      18         2
     111         3
     138         3
      53         3
      65         3
      31         5
      25         3
      63         3
      93         2
      62         4
      15         0
      78         3
      71         3
      79         0
      64         4
     122         0
      43         2
      66         1
      53         3
      99         3
      28         2
     110         0
     124         4
     123         2
     107         2
     105         5
      76         3
     114         4
      67         4
      13         2
      45         2
      72         1
     115         2
      71         3
      29         3
      79         1
      29         1
      30         3
      69         5
      13         1
      85         2
      81         3
      56         3
     102         1
      22         2
      84         3
     171         2
      99         2
     159         3
      32         4
     105         9
      23         2
      62         1
      94         2
      24         3
     110         1
      56         6
     100         2
      83         2
      93         3
      83         5
      51         3
     125         2
      42         2
      42         2
      22         2
      33         2
      52         5
      33         0
      89         4
      39         4
     131         3
     115         1
     111         2
      19         4
      42         4
      97         3
      28         3
     133         3
      54         3
      39         2
      23         2
      42         5
      16         1
      39         4
      48         3
     106         1
      45         4
      93         2
      35         2
      45         3
      43         4
     101         2
      55         3
      14         2
      68         2
      52         2
      44         3
      84         2
      73         4
      12         0
      27         3
      60         4
      84         2
      99         4
      76         3
      33         2
      20         2
      45         3
      54         2
      93         2
      38         3
      23         4
      16         0
      26         3
      68         4
      18         2
      98         1
      33         3
      42         3
      53         1
      29         4
      58         2
      45         3
      68         2
     137         0
      31         2
      45         2
      48         2
      48         2
      31         1
      95         6
      60         3
      16         2
     124         4
      71         3
     120         0
      51         2
      61         5
      63         3
      72         2
     135         4
     116         3
     134         3
      89         3
      67         3
      42         6
      74         4
      87         3
      20         1
      22         1
      43         2
      62         1
      14         2
      41         2
      51         2
      59         2
     111         2
      26         3
      71         2
      33         2
     132         1
     416         4
      44         4
      37         1
      81         2
      45         3
      47         4
      97         3
     134         1
      37         2
      22         4
     125         3
      16         2
     104         0
      44         3
      28         2
      88         3
      92         3
      62         1
      59         1
      33         2
      55         3
      53         5
      47         0
      69         4
      35         2
      52         3
     117         3
      22         3
      45         2
      80         1
      94         2
      49         2
      31         2
      78         3
      83         2
     265         5
      19         4
      66         0
      79         3
      30         6
      84         8
      49         3
      91         4
      49         3
      73         3
      35         1
      83         3
      25         1
      52         4
      43         3
      12         1
     246         5
      83         3
      80         4
     135         4
      58         2
     125         0
     104        10
      24         2
      88         1
     115         2
      39         3
      57         2
      27         3
      91         4
     154         6
      37         2
      70         3
      82         0
      51         3
      18         2
      50         3
      61         1
      77         4
      57         3
      78         2
      87         3
      43         4
      93         2
      65         3
      32         4
     391         2
      18         2
      24         2
      91         3
      25         2
      45         4
      55         3
      27         3
      41         2
     116         1
      77         6
     100         3
      88         4
      44         2
      16         2
      55         3
      64         1
      78         6
      35         3
      91         1
      46         4
      28         2
      94         2
      72         3
      86         2
      88         1
     149         4
      38         4
      18         3
      90         3
     103         1
      34         3
      20         1
      79         5
      31         6
      97         4
      23         1
      68         4
      45         3
      85         3
      40         1
      27         3
      75         3
      12         1
      99         5
      71         2
      64         0
      13         2
      56         3
      74         6
      44         0
      75         1
      68         2
      23         5
      19         2
      34         3
      72         3
      56         3
     141         0
      61         2
      97         2
      41         1
      23         3
      24         1
     131         7
      49         2
      93         2
      33         2
      94         0
      53         6
      71         5
      28         3
      86         4
      48         4
      13         0
      26         2
      47         2
      40         3
      63         3
     106         3
      65         2
      35         4
      34         1
      65         4
      36         3
      27         4
      25         2
      93         2
     109         4
      19         2
      47         5
      80         1
      86         2
      67         2
      58         3
      80         2
      43         2
     109         3
      60         3
      76         0
      54         2
      29         3
     104         2
      79         3
      63         3
      66         4
     115         5
      38         1
      18         4
      53         2
      44         1
     268         2
     114         5
      83         3
      23         2
     123         2
      33         3
      44         0
      97         1
      34         2
      64         1
      71         0
     114         0
      58         3
      87         2
      70         5
      32         3
      40         3
      78         4
      24         3
      51         3
      15         1
      47         3
      26         2
      42         2
      93         1
      77         0
      41         4
      39         4
      33         3
      81         1
      71         4
      78         2
      64         4
      96         5
     110         5
      13         1
      15         3
      67         5
      28         4
      97         2
      17         3
      17         0
      98         0
      77         0
      38         4
     105         1
      41         3
      72         1
      40         4
      90         3
      20         1
      26         0
      50         1
      18         2
      44         3
      92         1
      67         3
     103         1
      60         0
      20         1
      16         2
      21         2
      72         2
      23         3
      19         2
      28         3
      94         1
      66         0
      86         2
     101         3
      91         0
      93         2
      25         2
      62         3
      30         1
      57         3
      82         2
      58         2
      31         2
      62         4
      24         2
      74         3
      39         2
      58         2
      95         3
     102         1
      60         2
      72         3
      50         3
      68         6
}\dataForFigB
\paragraph{Computational cost management}
The primary hardware requirement for our memorization test is sufficient GPU memory to load the audited model and compute input gradients for a user-specified number of prompts. On an NVIDIA RTX A6000 GPU, assessing a single short ($\leq30$ tokens) sequence takes only a few minutes. The overall computational cost, therefore, is primarily a function of the total number of GCG runs required.

A possible concern is that this cost could become prohibitive. However, our methodology incorporates several features to ensure practical efficiency. First, as established in the previous section (analysis of the number of prefixes required), the vast majority of sequences require five or fewer prefixes to be classified as memorized. This result sets our heuristic budget of $\max(10, 2 \cdot P)$, ensuring that most sequences are assigned a modest, constant budget of 10 runs.
\begin{table}[h]
\centering
\caption{Computational cost statistics of the memorization test on the Pythia-6.9B model. Average allocated budgets and actual runs are shown for different datasets, highlighting efficiency via early stopping.}\label{tab:computational_cost}\vspace{-2mm}
\scalebox{0.9}{\begin{tabular}{lccc}
\toprule
\textbf{Dataset} & \textbf{Avg.\ Budget} & \textbf{Avg.\ Runs} & \textbf{Efficiency} \\
\midrule
Famous Quotes & 10.08 & 3.76 & 37.3\% \\
Paraphrased & 10.31 & 8.25 & 80.0\% \\
\bottomrule
\end{tabular}}

\end{table}

More importantly, the \textit{actual} computational cost is often significantly lower than the allocated budget due to early stopping. This efficiency is most apparent on highly memorized data. For instance, as shown in Table~\ref{tab:computational_cost} in the Pythia-6.9B model's test on the Famous Quotes dataset, the average allocated budget was 10.08 runs, but the average number of runs actually executed was only 3.76, which translates into an efficiency score of 37.3\% (actual as a percentage of the budget). In contrast, for the largely non-memorized Paraphrased Famous Quotes dataset, the average actual runs were nearly 80\% of the 10.31-run allocated budget. Even in this case, the framework's adaptability allows for further optimization. For datasets expected to be non-memorized, one can adopt a stricter strategy, such as flagging a sequence as non-memorized after only a few consecutive GCG failures to minimize computational cost.

\section{Conclusion}
\label{sec:conclusion}
We introduced the multi-prefix memorization framework, a principled approach to measuring how deeply content is embedded in LLMs by quantifying the diversity of eliciting prompts. Unlike prior methods focused on single-path elicitation, our framework emphasizes the multiplicity of access routes, capturing not just the presence of memorization but its robustness. We experimentally demonstrated that adversarial success rates offer a reliable signal for distinguishing memorized from non-memorized content, while also enabling practical cost control via early termination. Our findings reveal that memorization varies with data type, model scale, and instruction tuning, and is unevenly distributed across sequences. By exposing the structure of LLM memory, our framework lays the groundwork for more effective auditing and mitigation strategies, offering both a new metric and a deeper understanding of model behavior.

\end{document}